\documentclass[11pt,a4paper]{article}
\usepackage{times,latexsym}
\usepackage{url}
\usepackage[T1]{fontenc}
\usepackage{bbm}
    \usepackage[acceptedWithA]{tacl2018v2}
\usepackage[]{tacl2018v2}

\usepackage{xspace,mfirstuc,tabulary}

\newif\iftaclinstructions
\taclinstructionsfalse 
\iftaclinstructions
\renewcommand{\confidential}{}
\renewcommand{\anonsubtext}{(No author info supplied here, for consistency with
TACL-submission anonymization requirements)}
\newcommand{\instr}
\fi

\iftaclpubformat 

\else

\fi

\title{Context-aware Adversarial Training for Name Regularity Bias in\\ Named Entity Recognition}

\author{Abbas Ghaddar, Philippe Langlais\textsuperscript{\textdagger}, Ahmad Rashid and Mehdi Rezagholizadeh   \\
	Huawei Noah's Ark Lab, Montreal Research Center, Canada \\
	\textsuperscript{\textdagger}RALI/DIRO, Universit\'e de Montr\'eal, Canada \\
	{\tt abbas.ghaddar@huawei.com, felipe@iro.umontreal.ca}  \\
	{\tt ahmad.rashid@huawei.com, mehdi.rezagholizadeh@huawei.com}
	\\}

\hypersetup{draft}

\newcommand{\elmo}{ELMo}
\newcommand{\bert}{BERT}
\newcommand{\flair}{\textsc{Flair}}
\newcommand{\conll}{\textsc{CoNLL}}

\newcommand{\onto}{\textsc{OntoNotes}}

\newcommand{\per}{\textsc{per}}
\newcommand{\loc}{\textsc{loc}}
\newcommand{\org}{\textsc{org}}

\newcommand{\felipe}[2]{\textbf{\textcolor{purple}{#1}}}

\newcommand{\mightmention}[1]{}
\newcommand{\problem}[1]{\textcolor{red}{$\star$}}
\newcommand{\answer}[1]{\textcolor{blue}{$\#$}}
\newcommand{\todoreview}[1]{\textcolor{green}{$@$}}

\usepackage{subcaption}
\usepackage{rotating,csquotes}
\usepackage{xcolor}
\usepackage{multirow}
\usepackage{framed}
\usepackage{booktabs}
\usepackage{amsmath}
\usepackage{upgreek}
\usepackage{ulem}
\usepackage{amsfonts}

\newcommand{\nrb}{NRB}
\newcommand{\snrb}{\textsc{nrb}}
\newcommand{\fetest}{\textsc{test}}

\newcommand{\witness}{WTS}
\newcommand{\switness}{\textsc{wts}}

\usepackage[most]{tcolorbox}

\newtcbox{\mybox}[1][]{enhanced, colframe=blue, colback=blue!15, 
	frame style={opacity=0.25}, interior style={opacity=0.25}, 
	nobeforeafter, tcbox raise base, shrink tight, extrude by=1mm, #1}

\date{}

\begin{document}
\maketitle

\begin{abstract}
	In this work, we examine the ability of NER models to use contextual information when predicting  the type of an ambiguous entity. We introduce \nrb, a new testbed carefully designed to diagnose \textbf{N}ame \textbf{R}egularity \textbf{B}ias of NER models. Our results indicate that all state-of-the-art models we tested  show such a bias; \bert{} fine-tuned models significantly  outperforming feature-based (LSTM-CRF) ones on \nrb, despite having comparable (sometimes lower) performances on standard benchmarks. 
	
	To mitigate this bias, we propose a novel model-agnostic training method which adds~\textit{learnable} adversarial noise to some entity mentions, thus enforcing models to focus more strongly on the contextual signal, leading to significant gains on \nrb. Combining it with two other training strategies, data augmentation and parameter freezing, leads to further gains. 
\end{abstract}

%

\section{Introduction}

\noindent Recent advances in  language model pre-training~\cite{peters2018deep,devlin2018bert,liu2019roberta} have greatly improved the performance of many Natural Language Understanding (NLU) tasks. Yet, several studies~\cite{mccoy2019right,clark2019don, utama2020towards} revealed that state-of-the-art NLU models often make use of surface patterns in the data that do not generalize well.   Named-Entity Recognition (NER), a downstream task that consists in identifying textual mentions and classifying them into a predefined set of types, is no exception.

\begin{figure}[!ht]
	
	\begin{framed}
		\textbf{Gonzales, Louisiana}
		
		\underline{Gonzales}$^{\textcolor{blue}{LOC}}_{\textcolor{red}{PER}}$ is \mybox{a small city} in Ascension Parish, Louisiana.
		\vspace{2mm}
		
		\textbf{Obama, Fukui}
		
		\underline{Obama}$^{\textcolor{blue}{LOC}}_{\textcolor{red}{PER}}$ \mybox{is located} in far southwestern Fukui Prefecture.
		\vspace{2mm}
		
		\textbf{Patricia A. Madrid}
		
		\underline{Madrid}$^{\textcolor{blue}{PER}}_{\textcolor{red}{LOC}}$ \mybox{won her} first campaign in 1978 ..
		\vspace{-1mm} 
		
		\textbf{Asda Jayanama} 
		
		\underline{Asda}$^{\textcolor{blue}{PER}}_{\textcolor{red}{ORG}}$ \mybox{joined his brother}, Surapong ... 
	\end{framed}
	
	\caption{Examples extracted from Wikipedia (title in bold) that illustrate name regularity bias in NER. Entities of interest are underlined, gold types are in blue superscript, model predictions are in red subscript, and context information is highlighted in purple. Models employed in this study disregard contextual information and rely instead on some signal from the named-entity itself.}  
	
	\label{fig:nrb_samples}
\end{figure}

The  robustness of modern NER models has received considerable attention recently~\cite{mayhew2019ner,mayhew2019robust,agarwal2020entity,zeng2020counterfactual,berniercolborne-langlais:2020:LREC}. Name Regularity Bias~\cite{lin2020rigourous,agarwal2020interpretability,zeng2020counterfactual} in NER occurs when a model relies on a signal coming from the entity name, and disregards evidences within the local context. Figure~\ref{fig:nrb_samples} shows examples where state-of-the-art models~\cite{peters2018deep, akbik2018contextual, devlin2018bert} fail to exploit contextual information. For instance, the entity \textit{Gonzales} in the first sentence of the figure is wrongly recognized  as a \textit{person}, while the context clearly signals that it is a \textit{location} (city).

To better highlight this issue, we propose \nrb, a testbed designed to accurately diagnose name regularity bias of NER models by harvesting natural sentences from Wikipedia that contain challenging entities, such as those in Figure~\ref{fig:nrb_samples}. This is different from previous works that evaluate models on artificial data obtained by either randomizing~\cite{lin2020rigourous} or substituting entities by ones from a pre-defined list~\cite{agarwal2020entity}. \nrb{}  is compatible with any annotation scheme, and is intended to be used as an auxiliary validation set.

We conduct experiments with the \textit{feature-based} LSTM-CRF architecture~\cite{peters2018deep,akbik2018contextual} and the \bert{}~\cite{devlin2018bert} \textit{fine-tuning} approach trained on standard benchmarks. The best LSTM-based model we tested is able to correctly predict 38\% of the entities in \nrb. \bert{-based} models are performing much better (+37\%), even if they (slightly) underperform on in-domain development and test sets. This mismatch in performance between \nrb{} and standard benchmarks indicates that context awareness of models is not rewarded by existing benchmarks, thus justifying \nrb{} as an additional validation set.

We further propose a novel architecture-agnostic adversarial training procedure~\cite{miyato2016adversarial}
in which \textit{learnable} noise vectors are added to named-entity words, weakening their signal, thus encouraging the model to pay more attention to contextual information. Applying it to both  feature-based LSTM-CRF and fine-tuned \bert{} models leads to consistent gains on \nrb{} (+13 points) while maintaining the same level of performance on standard benchmarks.  

The remainder of the paper is organized as follows. We discuss  related works in Section~\ref{sec:Related Work}.  We describe how we built \nrb{} in Section~\ref{sec:nrb}, and its use in diagnosing named-entity bias of state-of-the-art models in Section~\ref{sec:diagnose}. In Section~\ref{sec:cure}, we present  a novel adversarial training method that we compare and combine with two simpler ones. We further analyze these training methods in Section~\ref{sec:further}, and conclude in Section~\ref{sec:conclusion}.

%

\section{Related Work}
\label{sec:Related Work}
\label{sec:related}

Robustness and out-of-distribution generalization has always been a persistent concern in deep learning applications such as computer vision~\cite{szegedy2013intriguing,recht2019imagenet}, speech processing~\cite{seltzer2013investigation,borgholt2020end}, and NLU~\cite{sogaard2013part,hendrycks17baseline,ghaddar2017winer,yaghoobzadeh2019robust,hendrycks2020pretrained}. One key challenge behind this issue in NLU is the tendency of models to quickly leverage surface form features and annotation artifacts~\cite{gururangan2018annotation}, which is often referred to as dataset biases~\cite{dasgupta2018evaluating,shah2020predictive}. We discuss  related works along two axes: diagnosis and mitigation. 

\subsection{Diagnosing Biais}

A growing number of  studies~\cite{zellers2018swag,poliak2018hypothesis,geva2019we,utama2020towards,sanh2020learning} are showing that NLU models rely heavily on spurious correlations between output labels and surface features (\textit{e.g.} keywords, lexical overlap),  impacting their generalization performance. Therefore, considerable attention has been paid to design diagnostic benchmarks where models relying on bias would perform poorly. For instance, HANS~\cite{mccoy2019right}, FEVER Symmetric~\cite{schuster2019towards}, and PAWS~\cite{zhang2019paws} are benchmarks that contain counterexamples to well-known biases in the training data of textual entailment~\cite{williams2017broad}, fact verification~\cite{thorne2018fever}, and paraphrase identification~\cite{wang2018glue} respectively.

Naturally, many entity names have a strong correlation with a single type (e.g. <Gonzales, PER> or <Madrid, LOC>). Recent works have noted that over-relying on entity name information negatively impacts NLU tasks. \citet{balasubramanian2020s} found that substituting named-entities in standard test sets of natural language inference, coreference resolution, and grammar error correction has a negative impact on those tasks. In political claims detection~\cite{pado2019sides}, \citet{dayanik2020masking} show that claims made by frequently occurring politicians in the training data are better recognized than those made by less frequent ones. 

Recently, \citet{zeng2020counterfactual} and \citet{agarwal2020interpretability} conducted two separate analyses on the decision making mechanism of NER models. Both works found that context tokens do contribute to system performance, but that entity names play a major role in driving high performances. \citet{agarwal2020entity} reported a performance drop in NER models when entities in standard test sets are substituted with other ones pulled from pre-defined lists. Concurrently, \citet{lin2020rigourous} conducted an empirical analysis on the robustness of NER models in the open domain scenario. They show that models are biased by strong entity name regularity, and train$\backslash$test overlap in standard benchmarks. They observe a drop in performance of 34\% when entity mentions are randomly replaced by other mentions. 

The aforementioned studies certainly demonstrate name regularity bias. Still, in many cases the entity mention is the only key to infer its type, as in "\textit{\underline{James} won the league}". Thus, randomly swapping entity names, as proposed by \newcite{lin2020rigourous},  typically introduces false positive examples,  which obscures observations.  Furthermore,  creating artificial word sequences introduces a mismatch between the pre-training and the fine-tuning phases of large-scale language models. 

NER is also challenging because of compounding factors such as entity boundary detection~\cite{zheng2019boundary}, rare words and emerging entities~\cite{strauss2016results}, document-level context~\cite{durrett2014joint}, capitalization mismatch~\cite{mayhew2019ner}, unbalance datasets~\cite{nguyen2020adaptive}, and domain shift~\cite{alvarado2015domain,augenstein2017generalisation}. It is unclear to us how randomizing mentions in a corpus, as proposed by \newcite{lin2020rigourous}, is interfering with these factors.

\nrb{} gathers genuine entities that appear in natural sentences extracted from Wikipedia. Examples are selected so that  entity boundaries are easy to identify, and their types can be inferred from the local context, thus avoiding compounding many factors responsible for lack of robustness. 

\subsection{Mitigating Bias}

The prevailing approach to address dataset biases consists in adjusting the training loss for biased examples. A number of recent studies \cite{clark2019don,belinkov2019adversarial,he2019unlearn,mahabadi2020end,utama2020mind} proposed to train
a shallow model that exploits manually designed biased features. A main model is then trained in an ensemble with this pre-trained model, in order to discourage the main model from adopting the naive strategy of the shallow one.

Adversarial training~\cite{miyato2016adversarial} is a regularization method which has been shown to improve not only robustness~\cite{ebrahimi2017hotflip,bekoulis2018adversarial}, but also generalization~\cite{cheng2019robust,zhu2019freelb} in NLU. It builds on the idea of adding adversarial examples~\cite{goodfellow2014explaining,fawzi2016robustness} to the training set, that is, small perturbations of the data that can change the prediction of a classifier. These perturbations for NLP tasks are done at the token embedding level and are norm bounded.  Typically, adversarial training algorithms can be defined as a \textit{minmax} optimization problem wherein the adversarial examples are generated to maximize the loss, while the model is trained to minimize it.  

\citet{belinkov2019adversarial} used adversarial training to mitigate the hypothesis-only bias in textual entailment models. \citet{clark2020learning} adversarially trained a low and a high capacity model in an ensemble in order to ensure that the latter model is focusing on patterns that should generalize better. \citet{dayanik2020masking} used an extra adversarial  loss in order to encourage a political claims detection model to learn more from samples with infrequent politician names. \citet{le2020adversarial} proposed an adversarial technique to filter-out biased examples from training material. Models trained on the filtered datasets show improved out-of-distribution performances on various computer vision and NLU tasks. 

Data augmentation is another strategy for enhancing robustness. It was successfully used in \cite{min2020syntactic} and \cite{moosavi2020improving} to improve textual entailment performances on the HANS benchmark. The former approach proposes to append original training sentences with their corresponding predicate-arguments triplets generated by a semantic role labelling tagger; while the latter generates new examples by applying syntactic transformations to the original training instances. 

\citet{zeng2020counterfactual} created  new examples by randomly replacing an entity by another one of the same type that occurs in the training data. New examples are considered valid if the type of the replaced entity is correctly predicted by a NER model trained on the original dataset. Similarly, \citet{dai2020analysis} explored different entity substitution techniques for data augmentation tailored to NER. Both  studies conclude that data augmentation techniques based on entity substitution improves the overall performances on low resource biomedical NER.

Studies discussed above have the potential to mitigate name regularity bias of NER models. Still, we are not aware of any dedicated work that shows it is so. In this work,  we propose ways  of mitigating name regularity bias for NER, including an elaborate adversarial method that enforces the model to capture more signal from the context. Our methods do not require an extra training stage, or to manually characterize biased features. They are therefore conceptually simpler, and can potentially be combined to any of the discussed techniques. Furthermore, our proposed methods are effective under both low and high resource settings.

%

\section{The \nrb{}  Benchmark}
\label{sec:nrb}

\nrb{} is a diagnosing testbed exclusively dedicated to name regularity bias in NER. To this end, it gathers named-entities that satisfy 4 criteria:

\begin{enumerate}
	
	\item Must be real-world entities within natural sentences $\rightarrow$ We select  sentences from Wikipedia articles. 
	
	\item Must be compatible with any annotation scheme  $\rightarrow$ We restrict our focus on the 3 most common types found in NER benchmarks:  \textit{person}, \textit{location}, and   \textit{organization}.
	
	\item Boundary detection (segmentation) should not be a bottleneck  $\rightarrow$ We only select single word entities that start with a capital letter.
	
	\item Supporting evidences of the type must be restricted to local context only (a window of 2 to 4 tokens) $\rightarrow$ We developed a primitive \textit{context-only} tagger to filter-out entities with no close-context signal.

\end{enumerate}

\begin{figure}[!th]
\resizebox{\columnwidth}{!}{
	\hskip-0.1cm\begin{tabular}{|ll|}
		\hline
		\textbf{Disambiguation page} &\\
		\multicolumn{2}{|l|}{\hspace*{5mm}\textit{Bromwich (disambiguation)}} \\
		\textbf{Query term}   & Bromwich \\ 
		& \\
		\textbf{Wikipedia article}  &  \textit{Kenny Bromwich} \\
		\textbf{Freebase type}  & \textit{PER} \\ 
		\textbf{Sentence} & \\
		\multicolumn{2}{|p{7.4cm}|}{\hspace*{5mm} Round 5 of the 2013 NRL season \underline{Bromwich} made his NRL debut for the Melbourne Storm}        \\  
		& \\
		\textbf{Taggers} & \\
		\multicolumn{2}{|l|}{\textbf{weak supervision}  \hspace*{14mm} \org{} (\textbf{confidence}: 0.97)} \\ 
		\multicolumn{2}{|l|}{\textbf{context-only} \hfill  \per: 0.58, \org: 0.30, \loc: 0.12} \\
		\hline							 				
	\end{tabular}
	}
	\caption{Selection of a sentence in  \nrb.}
	\label{fig:selection_criteria}

	
\end{figure}

The  strategy used to gather examples in \nrb{} is illustrated in Figure~\ref{fig:selection_criteria}. We first select Wikipedia articles that are listed in a disambiguation page. Disambiguation pages group different topics that could be referred to by the same query  term.\footnote{\url{https://en.wikipedia.org/wiki/Wikipedia:Manual_of_Style/Disambiguation_pages}.} The query term \textit{Bromwich}  in Figure~\ref{fig:selection_criteria} has its own disambiguation page that contains a link to the city of \textit{West Bromwich}, \textit{West Bromwich Albion Football Club}, and \textit{Kenny Bromwich} the rugby league player.

We associate each article in a disambiguation page to the entity type found in its corresponding Freebase page~\cite{bollacker2008freebase}, considering only  articles whose Freebase type can be mapped to a person, a location, or an organization. We assume that occurrences of the query term within the article are of this type. This assumption was found accurate in previous works on Wikipedia distant supervision for NER~\cite{ghaddar2016coreference,ghaddar2018lrec}. The sentence in our example is extracted from the \textit{Kenny Bromwich} article, whose Freebase type can be mapped to a person. Therefore, we assume \textit{Bromwich} in this sentence to be a person.

To decide whether a sentence containing a query term is worth being included in \nrb, we rely on two NER taggers. One is a popular NER system which provides a confidence score to each prediction, and which acts as a \textit{weak superviser}, the other is a \textit{context-only} tagger we designed specifically (see section~\ref{sec:implementation}) to detect entities with a strong signal from their local context. A sentence is selected if the query term is incorrectly labeled with high confidence (score $>$ 0.85) by the former tagger, while the latter one labels it correctly with high confidence (a gap of at least 0.25 in probability between the first and second predicted types). This is the case of the sentence in Figure~\ref{fig:selection_criteria} where  \textit{Bromwich} is incorrectly labeled as an organisation by the weak supervision tagger, however correctly labeled as a person by the  context-only tagger.

\subsection{Implementation}
\label{sec:implementation}
We used the Stanford CoreNLP~\cite{CORENLP} tagger as our weak supervision tagger and developed a simple yet efficient method to build a    context-only tagger. For this, we first applied the Stanford tagger to the entire Wikipedia dump  and replaced all entity mentions identified by their tag. Then, we train a 5-gram language model on the resulting corpus using kenLM~\cite{Heafield-kenlm}. Figure~\ref{fig:context_tagger} illustrates how this model is deployed as an entity tagger: the mention is replaced by an empty slot and the language model is queried for each type. We rank the tags using the perplexity score given by the model to the resulting sentences, then we normalize those  scores to get a probability distribution over types.

\begin{figure}[!ht]
	
	\begin{framed}
		\textit{\underline{Obama}} is located in far southwestern Fukui Prefecture.
		\\
		\textbf{$<$?$>$} is located in far southwestern Fukui Prefecture.
		\\
		\hspace*{1cm}   \{\textit{LOC}: 0.61, \textit{ORG}: 0.28, \textit{PER}: 0.11\}		
	\end{framed}
	
	
	\caption{Illustration of a language model used as a context-only tagger.}
	
	\label{fig:context_tagger}
\end{figure}

We downloaded the Wikipedia dump of June 2020, which contains 30k disambiguation pages. These pages contain links to 263k articles, where only 107k (40\%) of them have a type in Freebase that can be mapped to the 3 types of interest.  The Stanford tagger identified 440k entities that match the query term of the disambiguation pages.  The thresholds discussed previously were  chosen to select around 5000 of the most challenging examples in terms of name regularity bias. This figure aligns with the number of entities present in the test set of the well-studied \conll{} benchmark~\cite{tjong2003introduction}. 

We assessed  the annotation quality, by asking a human to filter out noisy examples. A sentence was removed if it contains an annotation error, or if  the type of the query term cannot be inferred from the local context.  Only 1.3\% of the examples where removed, which confirms  the accuracy of our automatic procedure. \nrb{} is  composed of 5275 examples, and each sentence contains a single annotation (see Figure~\ref{fig:nrb_samples} for examples). 
 
\begin{table*}[!th]
	\begin{center}
		\begin{tabular}{l|cccc|cccc}
			\toprule
			\multirow{2}{*}{Model} & \multicolumn{4}{c|}{\conll{}}& \multicolumn{4}{c}{\onto{}} \\
			& Dev & Test & \nrb{} & \witness{} & Dev & Test & \nrb{} & \witness{}  \\
			\midrule
			\multicolumn{9}{c}{\textit{Feature-based}}\\
			\midrule
			Flair-LSTM & - & \textbf{93.03} & 27.56 & \textbf{99.58}  & - & 89.06  & 33.67 & 93.98 \\
			ELMo-LSTM & 96.69 & 92.47 & 31.65 &  98.24 & 88.31 & 89.38 &  34.34  & 94.90 \\
			BERT-LSTM & 95.94 & 91.94 & 38.34 & 98.08 &  86.12 & 87.28  & 43.07 & 92.04 \\    
			\midrule
			
			\multicolumn{9}{c}{\textit{Fine-tuning}}\\
			\midrule
			BERT-base & 96.18  & 92.19 & 75.54 & 98.67 & 87.23 & 88.19 & 75.34  & 94.22 \\
			BERT-large & \textbf{96.90} & 92.86 & \textbf{75.55} & 98.51 & \textbf{89.26} & \textbf{89.93} & \textbf{75.41} & \textbf{95.06} \\
			\bottomrule
		\end{tabular}
	\end{center}
	\caption{Mention level F1 scores of  models on \conll{} and \onto, as well as on \nrb{} and \witness.}
	\label{tab:res_nothing}
	
\end{table*}

\subsection{Control Set (\witness)}

In addition to \nrb, we collected a set of domain control sentences --- called \witness{} for \textsc{Witness} --- that contain the very same query terms selected in \nrb, but which were correctly labeled by both the Stanford (score $>$ 0.85) and the context-only taggers. We selected examples with a small gap ($<$ 0.1) between the first and second ranked type assigned to the query term by the latter tagger. Thus, examples in \witness{} should be easy to tag. For example, because \textit{Obama} the Japanese city (see Figure~\ref{fig:context_tagger}) is selected among the query terms in \nrb, we added an instance of \textit{Obama} the president.

Performing poorly on such examples\footnote{That is, a system that fail to tag \textit{Obama} the president as a person.} indicates a domain shift between \nrb{} (Wikipedia) and whatever dataset a model is trained on (we call it the in-domain corpus).  \witness{} is composed of 5192 sentences that have also been manually checked.

%

\section{Diagnosing Bias}
\label{sec:diagnose}

\subsection{Data}

To be comparable with state-of-the-art models, we consider two standard benchmarks  for NER: \conll{-2003}~\cite{tjong2003introduction} and \onto{ 5.0}~\cite{pradhan_conll-2012_2012} which include 4 and 18 types of named-entities respectively. \onto{} is 4 times larger than \conll{}, and both benchmarks mainly cover the news domain. We run experiments on the official train/dev/test splits, and report mention-level F1 scores, following previous works. Since in \nrb, there is only one entity per sentence to annotate, a system is evaluated on its ability to correctly identify the boundaries of this entity and its type. When we train on \onto{} (18 types) and evaluate on \nrb{} (3 types), we perform type mapping  using the scheme of ~\newcite{augenstein2017generalisation}.

\subsection{Systems}
\label{sec:Baseline}

Following ~\cite{devlin2018bert}, we term all approaches that learn the encoder from scratch as \textit{feature-based}, as opposed to the ones that fine-tune a pre-trained model for the downstream task. We conduct experiments using 3 \textit{feature-based} and 2 \textit{fine-tuning} approaches for NER:

\begin{itemize}
	
	\item \textbf{Flair-LSTM}  An LSTM-CRF model that uses \flair{}~\cite{akbik2018contextual} contextualized embeddings as main features.
	
	\item \textbf{ELMo-LSTM} The LSTM-CRF tagging model of \newcite{peters2018deep} that uses \elmo{} contextualized embeddings at the input layer.
	
	\item \textbf{BERT-LSTM}  Similar to the previous model, but replacing \elmo{} by a representation gathered from the last four layers of  \bert.
	
	\item \textbf{BERT-base} The fine-tuning approach proposed by \newcite{devlin2018bert} using the \bert{-base} model. 
	
	\item \textbf{BERT-large} The fine-tuning approach using the \bert{-large} model. 
	
\end{itemize}

We used Flair-LSTM off-the-shelf,\footnote{https://github.com/flairNLP/flair} and re-implemented other approaches using the default settings proposed in the  respective papers. For our reimplementations, we used early stopping based on performance on the development set, and report average performance over 5 runs. For \bert{-based} solutions, we adopt spanBERT~\cite{joshi2019spanbert} as a backbone model since it was found by \citet{li2019unified} to perform better on NER.

\subsection{Results}

Table~\ref{tab:res_nothing} shows the mention level F1 score of the  systems considered. \flair{-LSTM} and \bert{-large} are the best performing models on in-domain test sets, the maximum gap with other models being 1.1 and 2.7 on \conll{} and \onto{} respectively. These figures are in line with previous works. What is more interesting is the performance on \nrb. Feature-based models do poorly, Flair-LSTM underperforms compared to other models (F1 score of 27.6 and 33.7 when trained  on \conll{} and \onto{} respectively). Fine-tuned \bert{} models clearly perform better (around 75), but far from in-domain results (92.9 and 89.9 on \conll{} and  \onto{} respectively). Domain shift is not a reason for those results, since the performances on \witness{} are rather high (92 or higher). Furthermore, we found that the boundary detection (segmentation) performance on \nrb{} is above 99.2\% across all settings. Since  errors made on \nrb{} are neither due to segmentation nor to domain shift, they must be imputed to  name regularity bias of models. 

 It is worth noting that \bert{-LSTM} outperforms \elmo{-LSTM} on \nrb, despite underperforming on in-domain test sets. This may be because \bert{} was pre-trained on Wikipedia (same domain of \nrb), while \elmo{} embeddings were trained on the One Billion Word corpus~\cite{chelba2013one}. Also, we  observe that switching from  \bert{-base} to \bert{-large}, or training on 4 times more data (\conll{} versus \onto{}) does not help on \nrb{}. This suggests  that name regularity bias is neither a data nor a model capacity issue.

\subsection{Feature-based vs. Fine-tuning}

In this section, we analyze reasons for the drastic superiority  of fined-tuned models on \nrb. First, the large gap between \bert{-LSTM} and \bert{-base} on \nrb{} suggests that this is not related to the representations being used at the input layer. 

Second, we tested several configurations of \elmo{-LSTM} where we scale up the number of LSTM layers and hidden units. We observed a degradation of performance on dev, test and \nrb{} sets, mostly due to over-parameterized models. We also trained 9, 6 and 4 layers \bert{-base} models,\footnote{We used early exit \cite{xin-etal-2020-deebert} at the $k^{th}$ layer.} and still noticed a large advantage of \bert{} models on \nrb.\footnote{The 4-layer model has 53M parameters and performs 52\% on \nrb.} This suggests that the higher capacity of \bert{} alone can not explain all the gains.
 
Third, since by design, evidences on the entity type in \nrb{} reside within the local context, it is unlikely that gains on this set come from the ability of Transformers~\cite{vaswani2017attention} to better handle long dependencies than LSTM~\cite{hochreiter1997long}. To further validate this statement, we fine-tuned \bert{} models with randomly initialized weights, except the embedding layer. We noticed that this time, the performances on \nrb{} fall into the same range of those of feature-based models, and a drastic decrease (12-15\%) on standard benchmarks. These observations are in keeping with results from \cite{hendrycks2020pretrained} on the out-of-distribution robustness of fine-tuning pre-trained transformers, and also confirms observations made by \cite{agarwal2020interpretability}.

From these analyses, we conclude that the Masked Language Model (MLM) objective~\cite{devlin2018bert} that the \bert{} models were pre-trained with is a key factor driving superior performances of the \textit{fine-tuned} models on \nrb{}. In most cases, the target word is  masked or randomly selected, therefore the model must rely on the context to predict the correct target, which is what a model should do to correctly predict the type of entities in \nrb. 
We think that in fine-tuning, training for a few epochs with a small learning rate, helps the model to preserve the \textit{contextual behaviour} induced by the MLM objective. 

Nevertheless, fine-tuned models recording  at best an F1 score of 75.6 on \nrb{} do show some name regularity bias, and fail to capture  useful local contextual information.

\begin{figure*}[t]
	\begin{center}
		\includegraphics[scale=0.35]{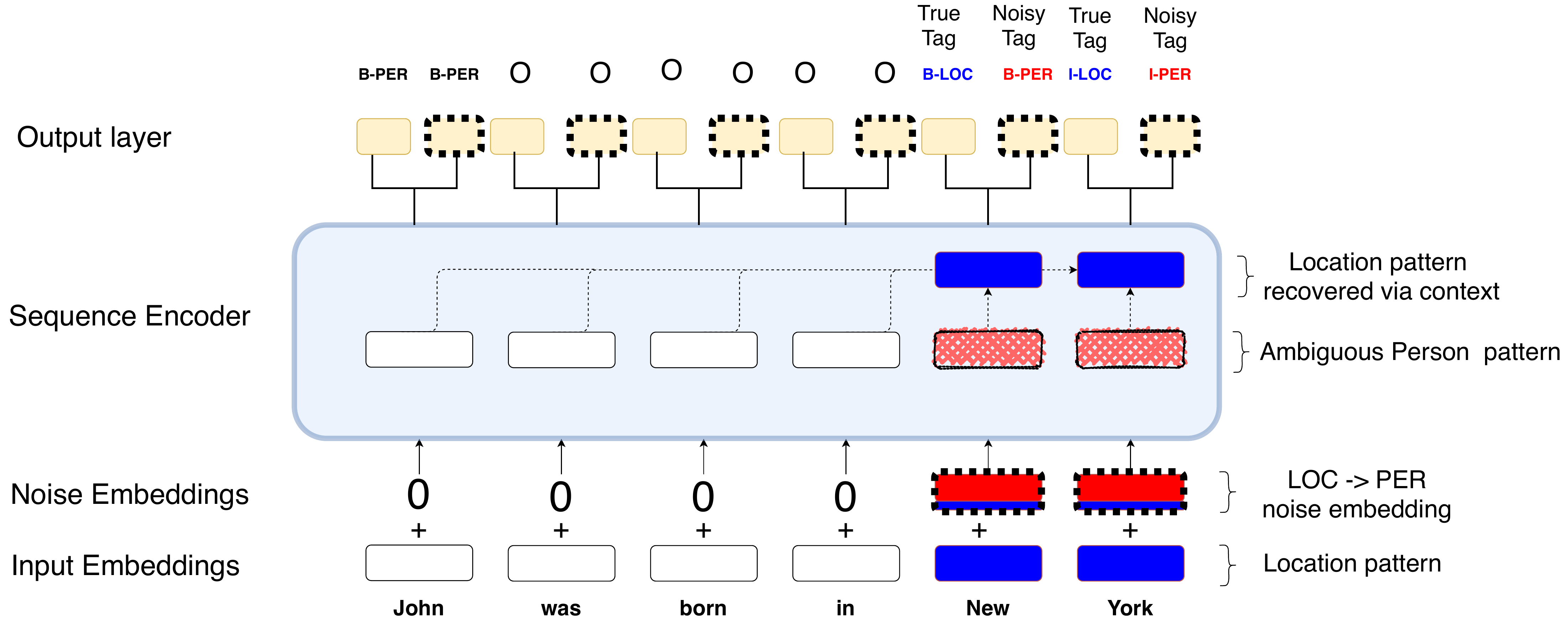} 
	\end{center}	
	\caption{Illustration of our adversarial method applied on the entity \textit{New York}. First, we generate a noisy type (PER), and then add a learnable noise embedding (\textsc{LOC$\rightarrow$PER}) to the input representation of that entity. This will make entity patterns (hashed rectangles) unreliable for the model, hence forcing it to collect evidences (dotted arrow) from the context. The noise embedding matrix and the noise label projection layer weights (dotted rectangle) are trained independently from the model parameters.}
	\label{fig:adv_method}
	
\end{figure*}

%
\section{Mitigating Bias}
\label{sec:cure}
 
In this section, we investigate training procedures that are designed to enhance the contextual awareness of a model, leading to a better performance on \nrb{} without impacting in-domain performance. These training procedures are not supposed to use any external data. In fact, \nrb{} is only used as a diagnosing corpus, once the model is trained. We propose 3 training procedures that  can be combined, two of them are architecture-agnostic, and one is specific to \textit{fine-tuning} \bert.   

\subsection{Entity Masking}

Inspired by the masking strategy applied during the pre-training phase of \bert{}, we propose a data augmentation approach that introduces a special \textsc{[MASK]} token in some of the training examples. Specifically, we search for entities in the training material that are preceded or followed by 3 non-entity words. This criterion applies to 35\% and 39\% of entities in the training data of \conll{} and \onto{} respectively. For each such entity, we create a new training example (new sentence) by replacing the entity by \textsc{[MASK]}, thus forcing the model to infer the type of masked tokens from the context. We call this procedure \textit{mask}.

\subsection{Parameter Freezing}

Another simple strategy, specific to \textit{fine-tuning} \bert{}, consists of freezing part of the network. More precisely, we freeze the bottom half of \bert{}, including the embedding layer. The intuition is to preserve part of the \textit{predicting-by-context} mechanism that \bert{} has acquired during the pre-training phase. This training procedure is expected to enforce the contextual ability of the model, thus adding to our analysis on the critical role of the MLM objective in pre-training \bert. We name this method \textit{freeze}.

\subsection{Adversarial Noise}

We propose an adversarial learning algorithm that makes entity type patterns in the input representation less reliable for the model, thus enforcing it to rely more aggressively on the context. To do so, we add a learnable adversarial noise vector (only) to the input representation of entities. We refer to this method as \textit{adv}.

Let $T=\{t_1, t_2, \ldots, t_K\}$ be a predefined set of types such as PER, LOC, and ORG in our case. 
Let $x=x_1, x_2, \ldots, x_n$ be the input sequence of length $n$,  $y = y_1, y_2, \ldots, y_n$ be the gold label sequence following the IOB\footnote{Naturally applies to other schemes, such as BILOU that \citet{ratinov2009design} found more informative.} tagging scheme, and $y^{\prime} = y^{\prime}_1, y^{\prime}_2, \ldots, y^{\prime}_n$ be a sequence obtained  by adding noise to $y$ at the mention-level, that is, by randomly replacing the type of mentions in $y$ with some noisy type sampled from $T$.
 
Let $\mathcal{Y}_{ij}(t)=y_i, \ldots, y_j$ be a mention of type $t\in T$, spanning the sequence of indices $i$ to $j$ in $y$.
We derive a noisy mention $\mathcal{Y'}_{ij}$ in $y'$ from $\mathcal{Y}_{ij}(t)$ as follows:
 
\[
\mathcal{Y}^{\prime}_{ij}= 
\begin{cases}
\mathcal{Y}_{ij}(t')             &{p\sim U(0,1)\leq \lambda}\\
& t' \sim \underset{\gamma\in T\setminus{\{t\}}}{\text{Cat}}( \gamma | \xi = \frac{1}{K-1} ) \\
\mathcal{Y}_{ij}(t) &  \text{otherwise} \\
\end{cases}
\]
 
 where $\lambda$ is a threshold parameter, $U(0,1)$ refers to the uniform distribution in the range [0,1],  Cat$(\gamma|\xi=\frac{1}{K-1})$ is the categorical distribution whose outcomes are equally likely with the probability of $\xi$, and the set $T\setminus{\{t\}}=\{t': t'\in T \wedge t' \neq t \}$ stands for the set $T$ excluding type $t$.

The above procedure only applies to the entities which are preceded or followed by 3 context words. For instance, in Figure~\ref{fig:adv_method}, we produce a noisy type for  \textit{New York} (PER), but not for \textit{John} ($p > \lambda$). Also, note that we generate a different sequence   $y^{\prime}$ from $y$  at each training epoch.

Next, we define a learnable noisy embedding matrix $E' \in \mathbb{R}^{m \times d}$ where $m=|T| \times (|T|-1)$ is the number of valid type switching possibilities,  and $d$ is the dimension of the input representations of $x$.  For each token with a noisy label, we add the corresponding noisy embedding to its input representation. {For other tokens, we simply add a zero vector of size $d$}. As depicted in Figure~\ref{fig:adv_method}, the noisy type of the entity \textit{New York} is \textsc{PER}, therefore we add the noise embedding at index $LOC \rightarrow{PER}$ to its input representation.

Then, the input representation of the sequence is fed to an encoder followed by an output layer, such as LSTM-CRF in \cite{peters2018deep}, or \bert{-Softmax} in~\cite{devlin2018bert}. First, we extend the aforementioned models by generating an extra logit $f^{\prime}$  using a projection layer parametrized by $W'$ and followed by a softmax function. As shown in Figure~\ref{fig:adv_method}, for each token the model produces two logits relative to the true and noisy tags. 
Then, we train the entire model to minimize two losses: $L_{true}(\theta)$ and $L_{noisy}(\theta')$, where $\theta$ is the original set of parameters and $\theta' = \{E^{\prime}, W^{\prime}\}$ is the extra set we added (dotted boxes in Figure~\ref{fig:adv_method}).
 $L_{true}(\theta)$ is the regular loss on the true tags, while $L_{noisy}(\theta')$ is the loss on the noisy tags defined as follows:

 \begin{equation}
    L_{\text{noisy}}(\theta') = \sum_{i=1}^n \mathbbm{1} (y'_i \neq y_i) \text{ CE}(f'_i,y'_i)
    \nonumber
\end{equation}

 where CE is the cross-entropy loss function. Both losses are minimized  using gradient descent. It is worth mentioning that $\lambda$ is  the only hyper-parameter of our \textit{adv} method. It  controls how often noisy embeddings are added during training. Higher values of $\lambda$ increase the amount of uncertainty around salient patterns in the input representation of entities, hence preventing the model from overfitting those patterns, and therefore pushing it to rely more on  context information. We tried values of $\lambda$ between $0.3$ and $0.9$, and found $\lambda=0.8$  to be the best one  based on \conll{} and \onto{} development sets.

\subsection{Results}

We trained models on \conll{} and \onto{}, and evaluated them on their respective \textsc{test} set.\footnote{Performances on \textsc{dev} show very similar trends.} Recall that \nrb{} and \witness{} are only used as  auxiliary diagnosing sets. Table~\ref{tab:adding_methods} shows the impact of our training methods when fine-tuning the \bert{-large} model (the one that performs best on \nrb).

First, we observe that each training method significantly improves the performance on \nrb{}. Adding adversarial noise is notably the best performing method on \nrb, with an additional gain of 10.5 and 10.4 F1 points over the respective baselines. On the other hand, we observe minor variations on in-domain test sets, as well as on \witness. The paired sample \textit{t-test}~\cite{cohen1996empirical} confirms that these variations are not statistically significant ($p > 0.05$). After all, the number of decisions that differ between the baseline and the best model on a given in-domain set is less than 20. 

\begin{table}[!th]

	\begin{center} 	 
	         
        \setlength{\tabcolsep}{4pt}

		\begin{tabular}{l|c l l| c l l}			\toprule
			
			\multirow{2}{*}{Method} & \multicolumn{3}{c|}{\conll}  & \multicolumn{3}{c}{\onto} \\
			& Test & \snrb & \switness  & Test & \snrb & \switness \\
			
			\midrule
			
			\bert{-lrg}  & 92.8 & 75.6 & \textbf{98.6} & 89.9 &  75.4  & 95.1\\ \hline 
			  \hspace{.5mm} +\underline{m}ask & \textbf{92.9} & 82.9 & 98.4  & 89.8 & 77.3 & \textbf{96.5} \\ 
			  \hspace{.5mm} +\underline{f}reeze & 92.7 & 83.1 & 98.4 & 89.9 & 79.8 & 96.0\\ 
			 \hspace{.5mm} +\underline{a}dv & 92.7 & 86.1 & 98.3 & \textbf{90.1} & 85.8 & 95.2\\ \hline
			 \hspace{.5mm} +f\&m & 92.8 & 85.5 & 97.8 & 89.9  & 80.6 & 95.9\\
			  \hspace{.5mm} +a\&m & 92.8 & 87.7  & 98.1 & 89.7  & 87.6  & 95.9\\ 
			  \hspace{.5mm} +a\&f & 92.7 & 88.4 & 98.2 & 90.0  & 88.1  & 95.7\\
			\hline
			 \hspace{.5mm} +a\&m\&f& 92.8 & \textbf{89.7} & 97.9  & 89.9 & \textbf{88.8} & 95.6\\ 
			
			\bottomrule
		\end{tabular}
	\end{center}
	\vspace{-0.5em}
	\caption{Impact of training methods on \bert{-large} models  fine-tuned on \conll{} or \onto.}
	\label{tab:adding_methods}
	
\end{table}

Second, we observe that combining methods always leads to improvements on \nrb{}; the best configuration being when we combine all 3 methods. It is interesting to note that combining training methods leads to a performance on \nrb{} which does not depend much on the training set used:  \conll{ }(89.7) and \onto{ }(88.8). This suggests that name regularity bias is a modelling issue, and not the effect of factors such as training data size, domain, or type granularity.

\begin{table}[!ht]
	
	\begin{center}
	\setlength{\tabcolsep}{4pt}
		\begin{tabular}{l|c ll| c ll}
			\toprule
			
			\multirow{2}{*}{Method} & \multicolumn{3}{c|}{\conll}  & \multicolumn{3}{c}{\onto} \\
			& Test & \snrb{} & \switness  & Test & \snrb{} & \switness  \\
			
			\midrule
			
			\textsc{E-LSTM}  & \textbf{92.5} & 31.7 & \textbf{98.2}  & \textbf{89.4}  & 34.3 & 94.9\\ \hline 
			\hspace{.5mm} +\underline{m}ask & 92.4 & 40.8 & 97.5  & 89.3  & 38.8 & \textbf{95.3} \\ 
			\hspace{.5mm} +\underline{a}dv & 92.4  & 42.4 & 97.8 & 89.4 &  40.7 & 95.0\\ 
			\hspace{.5mm} +a\&m & 92.4  &  \textbf{45.7} & 96.8 & 89.3 & \textbf{46.6} & 93.7 \\ 
			\bottomrule
		\end{tabular}
	\end{center}
	\vspace{-0.5em}
	\caption{Impact of training methods on the \elmo{-LSTM} trained on \conll{} or \onto.}
	\label{tab:elmo_lstm_methods}
\end{table}
 	 
In order to validate that our training methods are not specific to the \textit{fine-tuning} approach, we replicated the same experiments with the \elmo{-LSTM}. Table~\ref{tab:elmo_lstm_methods} shows the performances of the \textit{mask} and \textit{adv} procedures (the \textit{freeze} method does not apply here). The results are in line with those observed with \bert-{large}: significant gains on \nrb{} of 14 and 12 points for \conll{} and \onto{} models respectively, and no statistically significant changes on in-domain test sets. Again, combining training methods leads to systematic gains  on \nrb{} (13 points on average). Differently from fine-tuning \bert, we observe a slight drop in performance of 1.2\% on \witness{} when both methods are used.

The performance of \elmo{-LSTM} on \nrb{} does not rival with the one obtained by fine-tuning the \bert{-large} model, which confirms that \bert{} is a key factor to enhance robustness, even if  in-domain performance is not necessarily rewarded~\cite{mccoy2019right,hendrycks2020pretrained}.

%
\section{Analysis}
\label{sec:further}

So far, we have shown that state-of-the-art models do suffer from name regularity bias, and we proposed model-agnostic training methods which are able to mitigate this bias to some extent. In Section~\ref{sec:Attention Heads Analysis}, we provide further evidences that our training methods force the \bert{-large} model to better concentrate on contextual cues. In Section~\ref{sec:Random Permutations}, we replicate the evaluation protocol of \newcite{lin2020rigourous} in order to clear out the possibility that our training methods are only valid on \nrb.
Last, we perform extensive experiments on name regularity bias under low resource (Section~\ref{sec:Low Resource}) and multilingual (Section~\ref{sec:Multilingual}) settings.

\subsection{Attention Heads}
\label{sec:Attention Heads Analysis}

We leverage the attention map of \bert{} to better understand how our method enhances context encoding. To this end, we calculate the average number of attention heads that point to the entity mentions being predicted at each layer. We conduct this experiment on \nrb{} with the \bert{-large} model (24 layers with 16 attention heads at each layer) fine-tuned on \conll.  
 
 \begin{figure}[!ht]
	\begin{center}
		\resizebox{\columnwidth}{!}{	
			\includegraphics{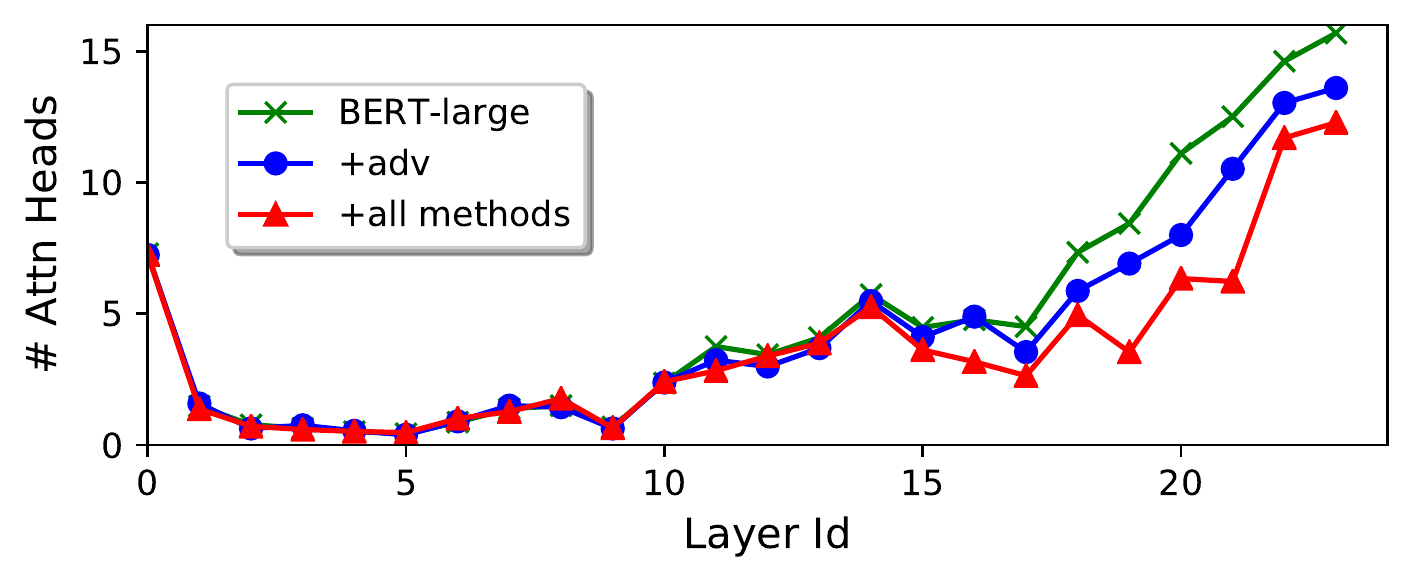}
		} 
	\end{center}	
	\caption{Average number of attention heads (y-axis) pointing to \nrb{} entity mentions at each layer (x-axis) of the \bert{-large} model fine-tuned on \conll.}
	\label{fig:att_head_ana}
	
\end{figure}

At each layer, we average the number of heads which have their highest attention weight (\textit{argmax}) pointing to the entity name.\footnote{We used the weights of the first sub-token since \nrb{} only contains single word entities.} Figure~\ref{fig:att_head_ana} shows the average number of attention heads that point to an entity mention in the \bert{-large} model fine-tuned without our methods, with the adversarial noise method (\textit{adv}), and with all three methods. 

We observe an increasing number of heads pointing to entity names when we get closer to the output layer: 
 at the bottom layers (left part of the figure) only a few heads are pointing to entity names, in contrast to the last 2 layers (right part) where almost all heads do so. This observation is inline with \newcite{jawahar2019does} who show that bottom and intermediate \bert{} layers mainly encode lexical and syntactic information, while top layers represent task-related  information.
Our training methods lead to less heads  at top layers pointing to entity mentions, suggesting  the model is focusing more on contextual information.

\subsection{Random Permutations}
\label{sec:Random Permutations}

Following the protocol described in \cite{lin2020rigourous}, we modified dev and test sets of standard benchmarks by randomly permuting dataset-wise mentions of entities, keeping the types untouched. For instance, the span of a specific mention of a person can be replaced by a span of a location, whenever it appears in the dataset. These randomized tests are highly challenging, as discussed in Section~\ref{sec:related}, since here the context is the only available clue to solve the task, and many false positive examples are introduced that way.

 \begin{table}[!ht]
	
	\begin{center}
		
		\begin{tabular}{l|l|l}
			\toprule
			
			Method & $\pi($dev$)$ & $\pi($test$)$ \\
			
			\midrule
			
			\bert{-large} & 23.45 & 25.46 \\ \hline 			
			\hspace{.5mm} +adv & 31.98  & 31.99   \\ 
			\hspace{.5mm} +adv\&mask & 35.02  & 34.09   \\ 
			\hspace{.5mm} +adv\&mask\&freeze & \textbf{40.39}   & \textbf{38.62}   \\
			
			\bottomrule
		\end{tabular}
		\vspace{-0.5em}
	\end{center}	
	\caption{F1 scores of \bert{-large} models fine-tuned on \conll{} and evaluated on randomly permuted versions of the  dev and test sets: $\pi($dev$)$ and $\pi($test$)$. }
	\label{tab:res_perm_conll}
\end{table}

Table~\ref{tab:res_perm_conll} shows the results of  the \bert{-large} model fine-tuned on \conll{} and evaluated on the permuted in-domain dev and test sets. F1 scores are much lower here, confirming this is a hard testbed, but they do provide evidences of the named-regularity bias of \bert. Our training methods improve the model F1 score by 17\% and 13\% on permuted dev and test sets respectively, an increase much inline with what we observed on \nrb.

\subsection{Low Resource Setting}
\label{sec:Low Resource}

Similarly to \cite{zhou2019dual,ding2020daga}, we simulate a low resource setting by randomly sampling tiny subsets of the training data. Since our focus is to measure the contextual learning ability of models, we first selected  sentences of \conll{} training data that contain at least one entity followed or preceded by 3 non-entity words. 

\begin{figure}[!ht]
	\begin{center}
		\resizebox{\columnwidth}{!}{	
			\includegraphics{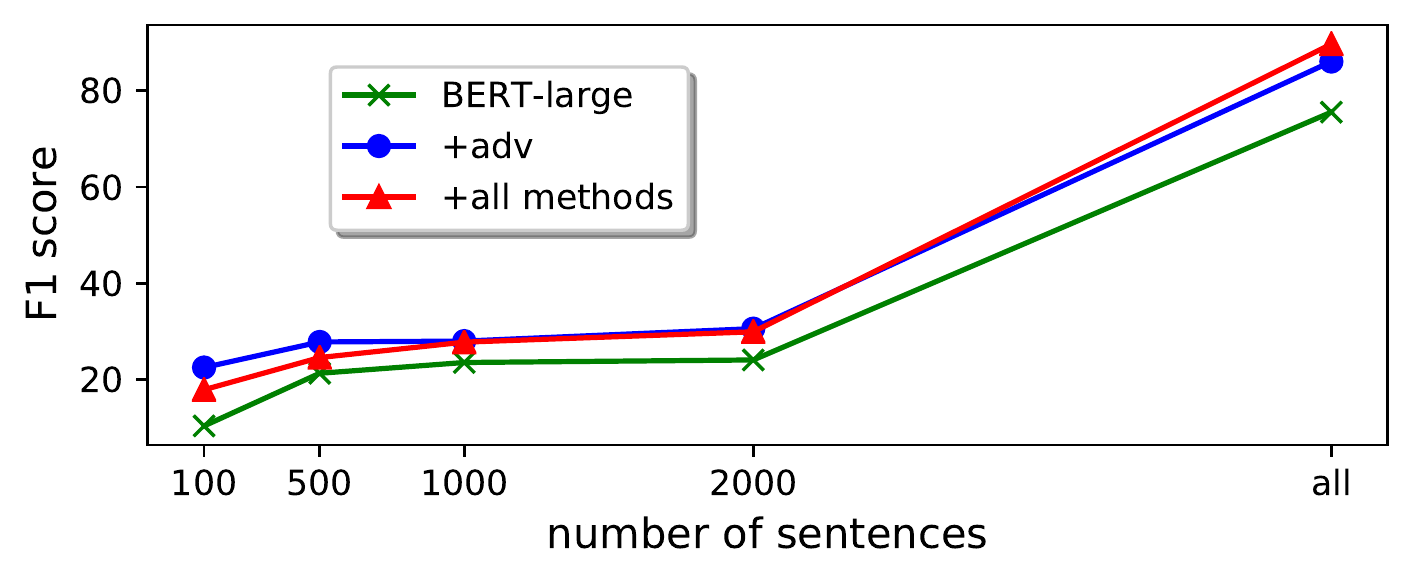}
		} 
	\end{center}	
	\caption{Performance on \nrb{} of \bert{-large} models as a function of the number of sentences used to fine-tune them. }
	\label{fig:lr_}
	
\end{figure} 

Then, we randomly sampled $k\in\{100, 500, 1000, 2000\}$ sentences\footnote{$\{0.7,3.5,7.1,14.3\}$\% of the training sentences.} with which we fine-tuned \bert{-large}. Figure~\ref{fig:lr_} shows the performance of the resulting models on \nrb. Expectedly, F1 scores of models fine-tuned with few examples are rather low on \nrb{} as well as on the in-domain test set. Not shown in Figure~\ref{fig:lr_}, fine-tuning on 100 and 2000 sentences leads to performance of 14\% and 45\% respectively on the \conll{} test set.  Nevertheless, we observe that our training methods, and \textit{adv} in particular, improve performances on \nrb{} even under extremely low resource settings. On \conll{} test and \witness{} sets, scores vary in a range of $\pm0.5$ and $\pm 0.7$ respectively when our methods are added to \bert{-large}.

\subsection{Multilingual Setting}
\label{sec:Multilingual}

\subsubsection{Experimental Protocol}

For in-domain data, we use the German, Spanish, and Dutch \conll{-2002}~\cite{tjong-kim-sang-2002-introduction} NER datasets. Those benchmarks --- also from the news domain --- come with a train/dev/test split, and the training material is comparable in size to the English \conll{} dataset. 
In addition, we experiment with four non \conll{} benchmarks: Finnish~\cite{luoma2020broad}, Danish \cite{hvingelby2020dane}, Croatian~\cite{Ljube2018croatain}, and Afrikaans~\cite{eiselen2016government} data. These corpora have more diversified text genres, yet mainly follow the \conll{} annotation scheme.\footnote{The Finnish data is tagged with EVENT, PRODUCT and DATE in addition to the \conll{} 4 classes.} Finnish and Afrikaans datasets have comparable size to English \conll{}, Danish is 60\% smaller, while the Croatian is twice larger. We use the provided  train/dev/test splits for Danish and Finnish, while we randomly split (80/10/10) the Croatian and Afrikaans datasets.

Since \nrb{} and \witness{} are in English, we designed a simple yet generic method for projecting them to another language. First, both test sets are translated to the target language using an online translation service. In order to ensure a high quality corpus, we eliminate a sentence if the BLEU score~\cite{papineni2002bleu} between the original (English) sentence and the back translated one is below 0.65.

\begin{table}[!th]
	
	\begin{center}
	
		
		\begin{tabular}{lll|lll}
			\toprule
			
			 & \nrb{} & \witness{} &  & \nrb{} & \witness{}\\ 
			
			\midrule
		    de & 37\% & 44\%  & fi & 53\%& 62\% \\ 
            es & 20\% & 22\% & da & 19\%& 24\%\\ 
            nl & 20\% & 24\% & hr  & 39\% & 48\%\\ 
            & & & af  & 26\% & 32\% \\
			\bottomrule
		\end{tabular}

		\vspace{-0.5em}
	\end{center}	
	\caption{Percentage of translated sentences from \nrb{} and \witness{}  discarded for each language. }
	\label{tab:multi_nrb}
\end{table}

\begin{table*}[!h]
	\begin{center}
	\setlength{\tabcolsep}{1.5pt}
	\resizebox{\textwidth}{!}{
		\begin{tabular}{l|ccc|ccc|ccc|ccc|ccc|ccc|ccc}
			\toprule
			\multirow{2}{*}{Model} & \multicolumn{3}{c|}{German} & \multicolumn{3}{c|}{Spanish} & \multicolumn{3}{c|}{Dutch} & \multicolumn{3}{c|}{Finnish} & \multicolumn{3}{c|}{Danish} & \multicolumn{3}{c|}{Croatian} 
			& \multicolumn{3}{c}{Afrikaans}\\
			 & \fetest & \snrb{} & \switness & \fetest & \snrb{} & \switness & \fetest & \snrb{} & \switness 
			 & \fetest & \snrb{} & \switness & \fetest & \snrb{} & \switness & \fetest & \snrb{} & \switness
			 & \fetest & \snrb{} & \switness \\  
			\midrule
			\multicolumn{22}{c}{\textit{Feature-based}}\\
			\midrule
			BERT-LSTM & 78.9 & 36.4 & 84.2 & 85.6 & 59.9 & 90.8 & 84.9 & 45.4 & 85.7 
			& 76.0 & 38.9 & 84.5 & 76.4 & 42.6 & 78.1 &  78.0 & 28.4 & 79.3 & 76.2 & 39.7 & 65.8\\
			\hspace{2mm}+adv & 78.2 & 44.1 & 82.8 & 85.0 & 65.8 & 90.2 & 84.3 & 57.8 & 83.5  
			& 75.1 & 52.9 & 81.0 & 75.4 & 47.2 & 76.9 & 77.5 & 35.2 & 75.5 &
			75.7 & 42.3 & 63.3\\
			\hspace{2mm}+adv\&mask  & 78.1	& 47.6 & 82.9 & 84.9 & 72.2 & 88.7 & 84.0 & 62.8 & 83.5  
			& 74.6 & 54.3 & 81.8 & 75.1 & 48.4 & 76.6 &  76.9 & 36.8 & 76.7 & 75.1 & 52.8 & 63.1\\
			\midrule
			
			\multicolumn{22}{c}{\textit{Fine-tuning}}\\
			\midrule
			BERT-base  & \textbf{83.8} & 64.0 & 93.3 & \textbf{88.0} & 72.3 & \textbf{93.9} & 91.8 & 56.1 & \textbf{92.0}  
			& \textbf{91.3} & 64.6 & 91.9  &  \textbf{83.6} & 56.6 & 86.2 & \textbf{89.7} & 54.7 & \textbf{95.6} & \textbf{80.4} & 54.3 & 91.6\\
			\hspace{2mm}+adv  & 83.7 & 68.9 & 93.6 & 87.9 & 75.9 & 93.9 & \textbf{91.9} & 58.3 & 91.8   
			& 90.2 & 66.4 & 92.5 & 82.7 & 58.4 & \textbf{86.5} &  89.5 & 57.9 & 95.5 & 79.7 & 60.2 & \textbf{92.1}\\
			\hspace{2mm}+a\&m\&f  & 83.2 & \textbf{73.3} & \textbf{94.0} & 87.4 & \textbf{81.6} & 93.7 & 91.2 &\textbf{63.6} & 91.0  
			& 89.8 & \textbf{67.4} & \textbf{92.7} & 82.3 & \textbf{63.1} & 85.4   & 88.8 & \textbf{59.6} & 94.9     & 79.4 & \textbf{64.2} & 91.6\\
			\bottomrule
		\end{tabular}
		}
	\end{center}
	\caption{Mention level F1 scores of 7 multilingual models trained on their respective training data, and tested on their respective in-domain test, \nrb{}, and \witness{} sets.}
	\label{tab:res_multi}
	
\end{table*}

Table~\ref{tab:multi_nrb} reports the percentage of discarded sentences for each language. While for the Finnish (fi), Croatian (hr) and German (de) languages we remove a large proportion of sentences, we found our translation approach more simple and systematic than generating an \nrb{} corpus from scratch for each language.  The latter approach depends on the robustness of the weak tagger, the number of Wikipedia articles and disambiguation pages per language, as well as the existence of type information. This is left as future work.

For experiments with \textit{fine-tuning}, we use language-specific BERT models\footnote{Language-specific models have been reported more accurate than multilingual ones in a monolingual setting~\cite{martin2019camembert,le2019flaubert,delobelle2020robbert,virtanen2019multilingual}.} for German~\cite{chan2020germans}, Spanish~\cite{canete2020spanish}, Dutch \cite{de2019bertje},  Finnish~\cite{virtanen2019multilingual}, Danish\footnote{\url{https://github.com/botxo/nordic_bert}}, Croatain~\cite{ulvcar2020finest}, while we use m\bert{}~\cite{devlin2018bert} for Afrikaans. 
 
For \textit{feature-based} approaches, we use the same architecture for \elmo{-LSTM}~\cite{peters2018deep} except that we replace English word embeddings by language-specific ones:   FastText~\cite{bojanowski2016enriching} for static representations, and the aforementioned \bert{-base} models for contextualized ones.

\subsubsection{Results}
 
Table~\ref{tab:res_multi} reports the performances on test, \nrb{}, and \witness{} sets for both \textit{feature-based} and \textit{fine-tuning} approaches with and without our training methods. We used the  hyper-parameters of the English \conll{} experiments with no further tuning. We selected the best performing models based on development sets score, and report average results on 5 runs.

Mainly due to implementation details and hyper-parameter settings, our fine-tuned \bert{-base} models perform better on the \conll{} test sets for German (83.8 vs. 80.4) and Dutch  (91.8 vs. 90.0) and slightly worse on Spanish (88.0 vs. 88.4) compared to the results reported in their respective \bert{} papers.

Consistent with the results obtained on English for \textit{feature-based} (Table~\ref{tab:res_nothing}) and \textit{fine-tuned} (Table~\ref{tab:elmo_lstm_methods}) models, the latter approach performs better on \nrb{}, although by a smaller margin compared to English (+37\%). More precisely, we observe a gain of +28\% and +26\% on German and Croatian respectively, and a gain ranging between 11\% and 15\% for other languages.

Nevertheless,  our training methods lead  to systematic and often drastic improvements on \nrb{} coupled with a statistically non significant overall decrease on in-domain test sets. They do however incur a slight but significant drop of around 2 F1 score points on  \witness{} for \textit{feature-based} models. Similar to what was previously observed, the best  scores on \nrb{} are obtained by \bert{} models when the training methods are combined. For the Dutch language, we observe  that once trained with our methods, the type of models used (feature-based versus \bert{} fine-tuned) leads to much less difference on \nrb.  \felipe{}{a bit short}

Altogether, these results demonstrate that name regularity bias is not specific to a particular language, even if its degree of severity varies from one language to another, and that the training methods proposed notably mitigate this bias.

%

\section{Conclusion}
\label{sec:conclusion}

In this work, we focused on the name regularity bias of NER models, a problem first discussed in \cite{lin2020rigourous}. We propose \nrb{,}  a benchmark we specifically designed to  diagnose such a bias. As opposed to  existing strategies devised to measure it, \nrb{} is composed of real sentences with easy to identify mentions. 

We show that current state-of-the-art models, perform from poorly (feature-based) to decently (fined-tuned \bert) on \nrb. In order to mitigate this bias,  we propose a novel adversarial training method based on adding some learnable noise vectors to entity words. These learnable vectors encourage the model to better incorporate contextual information. We demonstrate  that this approach greatly improves the contextual ability of existing models, and that it can be combined with other training methods we proposed. Significant gains are observed in both low-resource and multilingual settings. To foster research on NER robustness, we encourage others to report results on \nrb{} and \witness.\footnote{English and multilingual \nrb{} and \witness{} are available
at \url{http://rali.iro.umontreal.ca/rali/?q=en/wikipedia-nrb-ner}}

This study opens up new avenues of investigations. Conducting a large-scaled multilingual experiment, characterizing the name regularity bias of more diversified morphological language families is one of them, possibly leveraging massively multilingual resources such as WikiAnn \cite{pan2017cross}, Polyglot-NER~\cite{al2015polyglot}, or Universal Dependencies \cite{nivre2016universal}. We can also develop a more challenging \nrb{} by selecting sentences with multi-word entities.

Also, non-sequential labelling approaches for NER like the ones of \cite{li2019unified,yu2020named} have reported impressive results on both flat and nested NER. We plan to measure their bias on \nrb{} and study the benefits of applying our training methods to those approaches. Finally, we want to investigate whether our adversarial training method can be successfully applied to other NLP tasks.

\section{Acknowledgments}
 
We are grateful to the reviewers of this work for their constructive comments that greatly contributed to improving this paper.

\bibliography{references}
\bibliographystyle{acl_natbib}

\end{document}